\documentclass{vgtc}                          

\usepackage{mathptmx}
\usepackage{graphicx}
\usepackage{times}
\usepackage{xcolor}
\usepackage{amsmath}
\usepackage{todonotes}
\usepackage[colorlinks]{hyperref}
\usepackage{cleveref}
\usepackage{amssymb}
\usepackage[inline]{enumitem}

\usepackage[utf8]{inputenc}

\onlineid{0}

\vgtccategory{Research}

\vgtcinsertpkg

\title{Scene Graph Conditioning in Latent Diffusion}

\author{Frank Fundel\thanks{frank.fundel@uni-ulm.de}\\ %
     \scriptsize Universität Ulm %
}

\abstract{
Diffusion models excel in image generation but lack detailed semantic control using text prompts. Additional techniques have been developed to address this limitation. However, conditioning diffusion models solely on text-based descriptions is challenging due to ambiguity and lack of structure. In contrast, scene graphs offer a more precise representation of image content, making them superior for fine-grained control and accurate synthesis in image generation models. The amount of image and scene-graph data is sparse, which makes fine-tuning large diffusion models challenging. We propose multiple approaches to tackle this problem using ControlNet and Gated Self-Attention. We were able to show that using out proposed methods it is possible to generate images from scene graphs with much higher quality, outperforming previous methods. Our source code is publicly available on \url{https://github.com/FrankFundel/SGCond}
} 

\begin{document}

\maketitle


\section{Introduction}
Diffusion models \cite{diffusion, diffusion4img} have emerged as powerful image generation models, by gradually removing noise at each timestep. Recent advancements, such as Latent Diffusion \cite{latentdiffusion}, have improved the efficiency and computational cost of diffusion models. Latent Diffusion applies diffusion processes on a compressed latent space, achieved through a VQVAE-based autoencoder. Diffusion models exhibit great stability during training, avoid mode collapse, and generate images of exceptional quality. However, they are computationally expensive for sampling new images and lack a low-dimensional latent space compared to traditional VAEs, limiting conditional control. Nonetheless, diffusion models have proven to be highly effective in image generation, surpassing the performance of GANs in terms of image quality and stability \cite{diffusionbeatgans}.\\

Various techniques have been proposed to add more control to diffusion models. ControlNet \cite{controlnet} focuses on fine-tuning a copy of the model, while preserving the original knowledge using zero-convolutions. T2I-Adapter \cite{t2i-adapter} offers a lightweight model for fine-tuning only a smaller adapter model while leaving the diffusion model unchanged. GLIGEN \cite{gligen} introduces grounded text-to-image diffusion, combining text and grounding embeddings using Gated Self-Attention to enhance contextual understanding. These techniques showcase the advancements in controlling diffusion models.\\


While conditioning on text for image generation models has its limitations, scene graphs offer a more structured and unambiguous representation of image content. Text-based descriptions can be long, loosely structured, and subject to semantic ambiguity. In contrast, scene graphs provide a concise and precise encoding of object relationships and spatial arrangements within an image. These advantages make scene graphs a superior choice for conditioning image generation models when accurate synthesis and fine-grained control over visual content are crucial.\\

To address these challenges, different methods for utilizing scene graphs in image generation have been proposed. Some techniques rely on predicting scene layouts, which are coarse representations of the intended scenes \cite{sg2im, transformerbased}. Other methods condition on crops containing single objects corresponding to nodes in a scene graph \cite{pastegan}. Recently, works have been proposed which involve learning embeddings through masked contrastive pre-training to align image and graph features \cite{diffusionbased}. However, images that are conditioned on scene graphs are not yet comparable to images that are conditioned on other modalities such as bounding boxes, poses or depth maps.

\subsection*{Contribution}
Our contribution can be summarized with the following points:
\begin{itemize}
    \item We propose multiple approaches for effectively fine-tuning diffusion models using scene graphs
    \item We evaluated the most promising methods to some extend \textsuperscript{\ref{note:result}}
\end{itemize}

\section{Background}
\subsection{Image Generation}
Recently, image generation has acquired significant attention due to the remarkable advancements and capabilities demonstrated by deep learning models. These models have the potential to generate highly realistic and diverse images, often indistinguishable from those captured by cameras or created by humans \cite{diffusionbeatgans, gan4img}.

\subsubsection{Autoencoder}
Autoencoders are great at compressing and reconstructing images, while preserving their most prominent features. By encoding an image using a Convolutional Neural Network (CNN) into a compressed latent space representation and then decoding this representation back into an image, a simple and easy to train method for image generation is at hand - one can simply manipulate the latent vector and thus, induce the desired changes. But autoencoders also bring an interesting problem when it comes to image generation. Autoencoders map an input image directly into a single point in the latent space which results in a discrete distribution of points \cite{vae4img}. This makes interpolation between different latent representations difficult. Additionally, the distribution of points in the latent space is uncertain which results in poor sampling of novel images.

\subsubsection{GAN}
Generative Adversarial Networks (GANs) \cite{gan} solve the issue of non-uniform latent space by generating images directly from random uniform noise in the latent space. To train the generator network another network, the discriminator network, is trained simultaneously to decide if an image is from a set of example images or if it is generated. As the discriminator becomes better at deciding if an image is real or fake, the generator becomes better at generating new images. Because of the adversarial structure of the model, GANs are both very effective and can generate images almost indistinguishable from real images, but also difficult to train. Training is highly sensitive to hyperparameter selection and sometimes the model is unable to converge properly. Additionally, the generator can also learn to map every latent vector to the same few output images that are able to fool the discriminator. This phenomena is called mode collapse is one of the hardest problems GANs suffer from Ying et al. \cite{gan4img}.

\subsubsection{VAE}
Another approach that solves the non-uniform distribution problem of autoencoders are Variational Autoencoders (VAEs) \cite{vae}. Instead of directly encoding the input data into a single point in the latent space, the encoder network predicts the parameters $\sigma$ and $\mu$ of a probability distribution function. To reconstruct the image, a latent vector needs to be sampled from the distribution. If one took the predicted distribution parameters for reconstruction, the latent space would again be non-uniform. The uniformity comes from the sampling process. Rather than uniformly sampling from the predicted distribution function directly (which is not differentiable), a random variable $\epsilon$ is sampled from a uniform distribution e.g. standard Gaussian. A differentiable transformation using this random variable can then be defined (\autoref{eq:vae} compared to the original autoencoder defined by \autoref{eq:autoencoder}). To encourage the latent space distribution to be close to the prior distribution, Kullback-Leibler divergence is incorporated in the loss function \cite{vae4img}.

\begin{equation}
z = \mu + \exp(\log(\sigma^2) * \epsilon), \epsilon \sim \mathcal{N}(0, 1)
\label{eq:vae}
\end{equation}
\begin{equation}
z = \sigma(Wx+b)
\label{eq:autoencoder}
\end{equation}

This provides more certainty about the latent space distribution and prevents the formation of imprecise distributions that leads to inaccurate results \cite{vae4img}. VAEs are easier to train then GANs and are not susceptible to mode collapse, but commonly generate images with less quality. There are multiple causes for this issue \cite{msvae}. One cause is the use of Mean Squared Error (MSE) loss, because MSE is minimized by averaging all plausible outputs, which causes blurring. GANs mitigate this problem by using a discriminator network that only distinguishes real from fake, which is a more high-level goal \cite{pix2pix}. The other cause is that VAEs estimate the data distribution to be a normal Gaussian, where instead the real distribution could much more intricate.

This issue can be solved in VAEs by discretizing the latent space and applying the MSE not only on the pixel level but also on the latent level. Because the latent space is discrete, more fine-grained details are preserved \cite{vqvae2}. Vector Quantized VAEs (VQVAEs) \cite{vqvae} are designed for that task. Input images are encoded to closely match one of the learnable codebook vectors and the index of the closest (in terms of euclidean distance i.e. MSE) one is selected as the discrete representation. Due to continuous relaxation \cite{vqvae}, all codebook vectors are updated per image, thus a single codebook vector can be influenced by multiple similar images, allowing interpolation on a smooth latent space. Therefore, VQVAEs achieve an optimal combination of a discrete latent space, without the assumption of normal distributed data, while maintaining a smooth latent space. There are also applications of vector quantization in GANs (VQGAN\cite{VQGAN}), improving their performance by taming transformers. Previously transformers have shown better semantic quality because of the lack of inductive bias towards local interactions of CNNs, but worse resolution due to the quadratic complexity.

\subsubsection{Diffusion}
Latest image generation models rely on diffusion processes have shown great capabilities. Diffusion in machine learning was introduced by Sohl-Dickstein et al. in 2015 \cite{diffusion} as a generative model inspired by non-equilibrium statistical physics using Markov chains to gradually convert one distribution into another. This is done by adding a small amount of Gaussian noise to a sample at each timestep, resembling a Markov chain. In a Markov chain, each probability at each timestep only depends on the probability of the previous timestep and can be represented as a transition matrix \cite{markovchains}. This process of perturbing the input $x_0 \sim q(x_0)$ with noise over $T$ timesteps (\autoref{eq:pertube}) is called the forward process \cite{diffusion4img}. The Markov chain can be learned and reversed one step at a time by estimating the transpose transition matrix.\\

\begin{subequations}
\begin{align}
    \label{eq:pertube_a}
    q(x_t \mid x_{t-1}) &= \mathcal{N}(x_t ; \mu_t = \sqrt{1- \beta_t} x_{t-1}, \Sigma_t = \beta_t \mathbf{I}) \\
    \label{eq:pertube_b}
    q(x_{1:T} \mid x_0) &= \prod_{t=1}^T q(x_t \mid x_{t-1}) \\
    \label{eq:pertube_c}
    q(x_t \mid x_0) &= \mathcal{N}(x_t ; \sqrt{\bar{a}_t} x_{t-1}, (1-\bar{a}_t) \mathbf{I}) \\
    \label{eq:pertube_d}
    & = \sqrt{\bar{a}_t}x_0 + \epsilon \sqrt{1-\bar{a}_t}, \epsilon \sim \mathcal{N}(0, \mathbf{I})
\end{align}
\label{eq:pertube}
\end{subequations}
The first equation \autoref{eq:pertube_a} is the multivariate normal distribution, where $\mathbf{I}$ is the identity, indicating that each dimension has the same standard deviation $\beta_t$. The forward process is mathematically defined by the second equation \autoref{eq:pertube_b} as a product of all perturbations. To speed up this process, the product of Gaussians can also be defined as a single one with $\bar{a}_t = \prod_{s=1}^T (1-\beta_t)$, see \autoref{eq:pertube_c}.\\

The reverse distribution $q(x_{t-1} \mid x_t)$ is unknown and statistical estimates require computations involving the data distribution. However, a parameterized distribution $p_\theta(x_{t-1} \mid x_t)$ can be estimated by a neural network i.e. predicting $\mu_\theta(x_t, t)$ and $\Sigma_\theta(x_t, t)$. Ho et al. \cite{diffusion4img} claim that it is intractable to estimate the distribution without additionally conditioning on the original sample $x_0$, that is $q(x_{t-1} \mid x_t, x_0)$. The formula is re-parameterized to $\mu_\theta(x_t, t) = \mu_\theta(x_t(x_0, \epsilon), t)$. The final formula for $\mu_\theta$ shows that predicting the mean of the distribution is the same as predicting the noise $\epsilon$ at timestep $t$, thus the model is defined by $\epsilon_\theta(x_t, t)$. The function $q(x_t \mid x_0)$ for noising a sample is consequently also re-parameterized using $\epsilon$, see \autoref{eq:pertube_d}.\\

Hence, to train the model, each sample is produced by randomly drawing a data sample $x_0$, a timestep $t$, and noise $\epsilon$, which together give rise to a noised sample $x_t$. The objective becomes the MSE between the true noise and the predicted noise \cite{diffusion4img}. Specifically, after noising a sample, a U-Net is utilized to predict the noise, which can then be subtracted from the noisy sample. Because the amount of noise depends on the timestep, the model needs to know the current timestep. This is implemented using position embedding, introduced by Vaswani et al. \cite{AIAYN}. Different variance schedules $\beta_1, ..., \beta_T$ can be used e.g. linear, quadratic or cosine. To generate new samples, Gaussian noise $x_T$ and repeatedly predicting $x{t-1}$ from $x_t$. That is, the noise $\epsilon$ is predicted by the model and subtracted from the sample to obtain an estimate of $x_0$, then some noise is added back to the sample to get $x_{t-1}$. This is done iteratively until $t = 0$, generating better and better approximations of $x_0$. Song et al. \cite{DDIM} later proposed the Denoising Diffusion Implicit Model (DDIM), which is a diffusion model variant, that generalizes it to a non-Markovian process, while maintaining the same training procedure. This makes sampling much faster, while introducing a parameter $\eta$ to control the determinism of the model.\\

Training diffusion models is very computationally expensive because of the gradient computations on the high-dimensional pixel space of RGB images. And also during inference diffusion models are quite demanding in terms of compute resources, due to the sequential reverse process. Consequently, the training of such models is currently limited to a select minority within the field and furthermore, leaves a substantial carbon footprint \cite{latentdiffusion}. To address these issues, Rombach et al. introduce Latent Diffusion \cite{latentdiffusion}, where the two diffusion processes are applied not on the pixel space but on a compressed latent space. This is achieved by, first, training an autoencoder to provide a lower-dimensional latent space, and second, training a diffusion model on the compressed samples. Good compression is not necessary, because the only objective of the latent space is more efficient diffusion, thus training the autoencoder is much simpler. In particular, the authors utilize a variation of a VQGAN \cite{VQGAN} as the autoencoder and, but with the quantization layer in the decoder instead of the encoder. The goal of the VQGAN is to learn a latent space that is visually equivalent from the image space, while also providing a notable decrease in computational complexity. The U-Net for the noise prediction is a time-conditional U-Net, mainly comprised of 2D-convolutional layers. Sampling is done using the DDIM method. Their method achieve state-of-the-art results while requiring substantially less computational resources.\\

Diffusion models have shown to generate images with extremely high quality, even beating GANs \cite{diffusionbeatgans}. They are more stable during training and do not suffer from mode collapse. But sampling new images is quite expensive in terms of time and compute. Additionally, diffusion models lack a low-dimensional latent space, whereas standard VAEs typically have access to a low-dimensional latent space, which allows conditional control of the generation \cite{lowlatent} but limits their overall performance \cite{latentdiffusion}.\\

Summarizing, GANs provide fast sampling and high quality samples, whereas VAEs on the other hand provide a much higher diversity (mode coverage) as well as fast sampling speed. Diffusion models sampling speed is much slower, but come with high image quality and diversity.

\subsection{Image-Text Embeddings}
\label{sec:CLIP}
Aligning images with text is the process of accurately matching textual descriptions with their corresponding images. This alignment allows for a coherent and meaningful connection between the visual content of an image and its textual information. CLIP \cite{CLIP} is the most important work in this area. CLIP is a large transformer model trained from scratch on 400M image-text pairs scraped from the internet. The model consists of an image encoder (ViT \cite{ViT}) and a text encoder (Transformer \cite{transformer}). For efficiency, instead of predicting each exact word, the goal is to predict which text as whole belongs to an image. For that, the cosine similarity over a minibatch of correct image-text pairs is maximized during training, while minimizing the cosine similarity of the incorrect pairs. This is also known as contrastive learning, where a model learns without predicting labels or text, but instead learns a representation in which similar samples are close together, while dissimilar ones are far apart. CLIP provides zero-shot capabilities that are even competitive to supervised models.

\subsection{Conditional Image Generation}
By incorporating additional information into the image generation process, such as text, or other images, conditional image generation models enable some control over the generated output. Pix2Pix \cite{pix2pix}, proposed in 2016, introduced a conditional image-to-image translation framework based on GANs, with a U-Net as the generator and a specific loss function to learn from paired images. This model demonstrated impressive results in various tasks, such as generating realistic street scenes from semantic labels, transforming day-time images into night-time equivalents, or converting sketches into photorealistic images. CycleGAN \cite{cyclegan}, introduced in 2017, revolutionized conditional image generation by enabling the transformation of images from one domain to another without the need for paired training data. It employed a cycle consistency loss, which ensured that the transformation from one domain to another and back again would reconstruct the original image. One of the earliest text conditioned image generation networks was introduced in 2016 by Reed et al. \cite{textcond} by concatenating the embedded text to the input noise of the generator and to the features of the discriminator of a GAN.\\

Conditioning mechanisms in diffusion models use various forms of conditions, such as labels, classifiers, texts, images, semantic maps, and graphs. There are four main types of conditioning mechanisms: concatenation, gradient-based, cross-attention, and adaptive layer normalization (adaLN). Concatenation involves combining conditioning information with the input noise or intermediate denoised targets during the sampling process. For example, text embeddings \cite{latentdiffusion} and semantic feature maps \cite{semanticmap} can be used as guidance using this technique. Gradient-based mechanism incorporates gradients from a classifier into the diffusion sampling process \cite{diffusionbeatgans}. This allows for controllable generation, such as guiding the diffusion process towards a specific class label. Cross-attention \cite{AIAYN} is effective for training attention-based multi-modal models. In diffusion models, it is usually applied to intermediate layers of the denoising U-Net \cite{latentdiffusion}. Adaptive layer normalization (adaLN) replaces standard layer norm layers in transformer-based diffusion models with adaptive layer normalization. Instead of directly learning scale and shift parameters, adaLN regresses them from the sum of the time and condition embeddings \cite{adaln}, thus shifting the distribution. A variety of conditional diffusion models have already been proposed.

\subsubsection{Conditional Diffusion and Guided Diffusion}
Diffusion models try to predict the noise $\epsilon$ at timestep $t$, which is also equivalent to predicting the score function $\nabla_x \log p_t(x)$ \cite{scorebased}. This process can be conditioned on an arbitrary class $y$, resulting in the conditional score function $\nabla_x \log p_t(x \mid y)$. Traditional class-conditional diffusion models try to model the conditional distribution $p(x \mid y)$ during training, but performance can be improved when reformulating the score function to $\nabla_x \log p_t(x) + \nabla_x \log p_t(y \mid x)$ by using Bayes' rule. Now it can be seen that $\nabla_x \log p_t(y \mid x)$ is, in fact, what a classifier does. Dhariwal et al. \cite{diffusionbeatgans} showed that the conditional distribution can be approximated with the unconditional distribution with its mean shifted by $\Sigma \nabla_x \log p_t(y \mid x)$. The formula for the shifted mean is given by
\begin{equation}
    \hat{\mu}_\theta\left(x_t \mid y\right)=\mu_\theta\left(x_t \mid y\right)+s \cdot \Sigma_\theta\left(x_t \mid y\right) \nabla_{x_t} \log p_\phi\left(y \mid x_t\right)
\end{equation}
, where $\theta$ and $\phi$ are the parameters of two different models, and $s$ is the guidance scale. Now, instead of training a conditional diffusion model $p(x \mid y)$, a traditional pre-trained diffusion model can be guided using the gradients of a classifier trained on noisy images.

\subsubsection{Classifier-Free Guidance}
Training a classifier on extremely noisy images is quite difficult and adds another level of complexity to the image generation process. Additionally, during sampling, the use of a classifier on perturbed images can bee interpreted as adversarial attacks and thus limits the performance. In classifier-free guidance \cite{classifierfree}, a conditional diffusion model is utilized, denoted as $p_{\theta}(x_t \mid y)$. However, instead of using a specific label $y$, a null label $\varnothing$ is introduced with a fixed probability during training (usually 10\% of the time). The resulting model can serve as either a conditional model or an unconditional model,  depending on the presence or absence of a conditioning signal. During sampling, the two cases $p_{\theta}(x_t|y)$ and $p_{\theta}(x_t \mid \varnothing)$ can be used to guide the denoising process by:
\begin{equation}
    \hat{p}_\theta\left(x_t \mid y\right)=p_\theta\left(x_t \mid \emptyset\right)+s \cdot\left(p_\theta\left(x_t \mid y\right)-p_\theta\left(x_t \mid \emptyset\right)\right)
\end{equation}
By employing this method, the model does not depend on an additional classifier, but rather leverages its own knowledge. Finally, conditioning on free text using class-conditioned diffusion is difficult, as it is not trivial to build a classifier for that task.

\subsubsection{Conditioning on Text}
Text is a powerful medium for conveying rich and detailed information. By utilizing text as a modality for image generation, one can leverage the descriptive nature of language to convey specific attributes, spatial relationships, and contextual information that can enhance the fidelity and accuracy of the generated images. Moreover, text offers a flexible and accessible means of input for users, as it can be easily generated, manipulated, and understood.\\

Text-condition in state-of-the-art models is often done using CLIP. Perhaps, the biggest breakthrough in text-to-image was made by Ramesh et al. with unCLIP \cite{unclip} (also known as DALLE-2), introducing a two-stage model that generates CLIP image embeddings based on text captions and decodes high-quality images conditioned on those embeddings using diffusion. Thereby, enabling language-guided image manipulations in a computationally efficient manner and pushing the boundaries of text-to-image generation.\\

Prior to unCLIP, the authors of GLIDE \cite{GLIDE} compared two variants of text-conditioned diffusion: CLIP guidance and classifier-free guidance, where the latter performed better. For the first method, instead of using gradients from a classifier to shift the mean at each step in the sampling process, the gradients of the dot product of the image $x_t$ and caption $c$ encodings w.r.t. the image are used (\autoref{eq:clipcond}). Similar to classifier guidance, CLIP has to be trained on noisy images and the diffusion model itself does not have to be trained specifically.
\begin{equation}
    \hat{\mu}_\theta\left(x_t \mid c\right)=\mu_\theta\left(x_t \mid c\right)+s \cdot \Sigma_\theta\left(x_t \mid c\right) \nabla_{x_t}\left(f\left(x_t\right) \cdot g(c)\right)
\label{eq:clipcond}
\end{equation}
For the second method, the authors utilize a Transformer \cite{AIAYN} model to create text embeddings, which are then used as conditioning vectors for the diffusion model. Additionally, the feature vectors of the last layer of the Transformer are concatenated on to each attention layer of the denoising U-Net.\\

unCLIP employs the same architecture as GLIDE, but adding four extra CLIP tokens to the text embeddings of GLIDE and adding CLIP embeddings to the existing timestep embedding. The authors implement classifier-free guidance to enhance the sampling quality. unCLIP uses a prior network to map the CLIP text embeddings to the corresponding image embeddings. Although CLIP aims to minimize text and image embeddings for ranking purposes, its training objective does not explicitly focus on aligning modalities in the latent space. Instead, the model is trained to ensure that matching examples, where the text and image correspond to each other, are closer together in the embedding space compared to non-matching examples. This ranking objective allows CLIP to associate relevant text and image pairs effectively, enabling it to perform tasks such as image classification or zero-shot image generation based on textual prompts, see \autoref{sec:CLIP}. However, the specific spatial alignment of the modalities in the latent space is not the primary goal of CLIP's training process (also known as modality gap \cite{modalitygap}). The authors of unCLIP state that it is indeed possible to generate images directly on the CLIP text embeddings or the text tokens itself (e.g. GLIDE \cite{GLIDE}), but with worse results. unCLIP enables three different kinds of image manipulations. First, one can interpolate between two different images by interpolating between the CLIP embeddings. Second, by subtracting CLIP embeddings, one can remove aspects of an image. Third, one can generate variations of an image by choosing $\eta > 0$ (where $\eta = 0$ makes the model deterministic).\\

Latent Diffusion \cite{latentdiffusion} also incorporates classifier-free guidance, but uses a domain-specific (e.g. CLIP for text) encoder to obtain $y$ and guides the sampling using cross-attention in the denoising U-Net.

\subsubsection{Conditioning on other Modalities}
Following the successful development of text-to-image diffusion models, researchers have explored methods to incorporate additional modalities such as semantic maps, depth maps, and other images. They can be encoded and used as guidance or used as an initial image and guided with text. In Stable Diffusion (e.g. diffusers \cite{img2img}), when used as an initial image, noise is added to the input. The strength parameter, in essence, determines the extent of transformation applied to the reference image, with higher strength values leading to greater noise addition. However, when conditioning the model on an initial image, depth map, or other modalities, it becomes necessary to fine-tune the model accordingly for the specific modality.\\

While fine-tuning diffusion models can be beneficial for incorporating specific modalities or addressing certain tasks, there are also some disadvantages associated with this approach. First, the amount of data in task-specific domains is often fairly limited, which can lead to overfitting and less generalization when fine-tuning large models. And second, fine-tuning large diffusion models requires a substantial amount of computational resources, thereby restricting the potential for personalization \cite{controlnet}.

\subsection{Scene Graphs}
Scene graphs are a special kind of graph used in computer vision and provide a powerful structural representation of an image. They depict scenes as directed graphs, with objects as nodes and relationships between objects as edges \cite{sg2im}. These graphs can be predicted from images and utilized for tasks such as image retrieval or image captioning \cite{pastegan}. Scene graphs provide a structured and versatile means of describing multiple objects and their intricate relationships within an image. Moreover, they enable generative models to reason over both objects and their relations when generating images. 

\section{Related Work}
\subsection{Efficient Fine-Tuning}
Fine-tuning large diffusion models comes with several challenges, mainly data and compute resource limitations. Thus, using a smaller network to influence a large one can be beneficial. NovelAI \cite{NovelAI} introduced a method on how to build and train such a HyperNetwork for Stable Diffusion. Their method consists of a small number of linear layers that are connected to the cross-attention module of the denoising U-Net, altering the key and values from the text guidance. Throughout the training process, only the HyperNetwork is updated, while the diffusion model remains unchanged. Due to the compact size of the HyperNetwork, training is fast and requires minimal compute resources.\\

Another method, LoRA \cite{lora}, works by training two smaller matrices $A \in \mathbb{R}^{r \times k}$ and $B \in \mathbb{R}^{d \times r}$ (rank decomposition matrices), which dot-product $\delta W = BA$ has the same size as the cross-attention weights $W_0 \in \mathbb{R}^{d \times k}$ of the denoising U-Net. Their size can be controlled by $r << \min(d, k)$. The resulting matrix $\delta W$ is then added to the initial weight matrix $W_0$. During training the diffusion model remains frozen and only the smaller matrices $A$ and $B$ are updated.\\

ControlNet (Feb 2023) is a fine-tuning method that aims to preserve the original knowledge of the larger model, while preventing overfitting when the fine-tuning dataset is small. ControlNet can be used as a robust method to fine-tuning large diffusion models. This is achieved by freezing the original layers of the denoising U-Net and creating trainable copies of them. Zero-convolution is applied to the additional condition, which is then added to the input for the trainable copy. Afterwards, zero-convolution is applied again to the output of the trainable copy, which is then added to the output of the frozen layer. Zero-convolution, hereby stands for a standard $1 \times 1$ convolution layer, where the weights are initialized with zeros. This turns the ControlNet into the identity function in the beginning of the training, thus preserving the functionality of each block.\\

T2I-Adapter (Mar 2023) \cite{t2i-adapter} proposes a light-weight model consisting of four feature extraction blocks and three downsample blocks, where a feature extraction block consists of one convolution and two residual blocks. The size of the features extracted in each feature extraction block align with the output size of each layer in the denoising U-Net, allowing them to be added. The diffusion model is not updated during training, only the adapter is. T2I-Adapter also allows multiple conditions, by adding the features of multiple adapters.\\

GLIGEN (Apr 2023) \cite{gligen} introduces grounded text-to-image diffusion by combining text embeddings $c$ with certain grounding embeddings $e$. These grounding embeddings can be generated from a variety of input modalities such as depth maps, pose estimations, semantic maps etc. but also from a combination of text and corresponding bounding boxes. To incorporate the new conditional information, Gated Self-Attention layers are placed between each Cross-Attention layer. In Gated Self-Attention, the visual tokens from the diffusion process and the conditioning tokens are combined and Self-Attention is applied. Afterwards, only the visual tokens are selected and gated by multiplying it with a learned parameter (initialized with 0). During training, all other layers are fixed and only the new gated cross-attention layers are updated.\\

DiffFit (May 2023) \cite{difffit} drastically reduces trainable parameters while fine-tuning large diffusion models by freezing all parameters except for bias, Layer-Normalization and class embedding. Additionally a trainable scaling-term $\gamma$ is inserted (initialized to 1) and multiplied to each encoder block. This method can also be combined with other fine-tuning models such as ControlNet, freezing the trainable copy.

\begin{figure*}
\centering
\includegraphics[width=\textwidth]{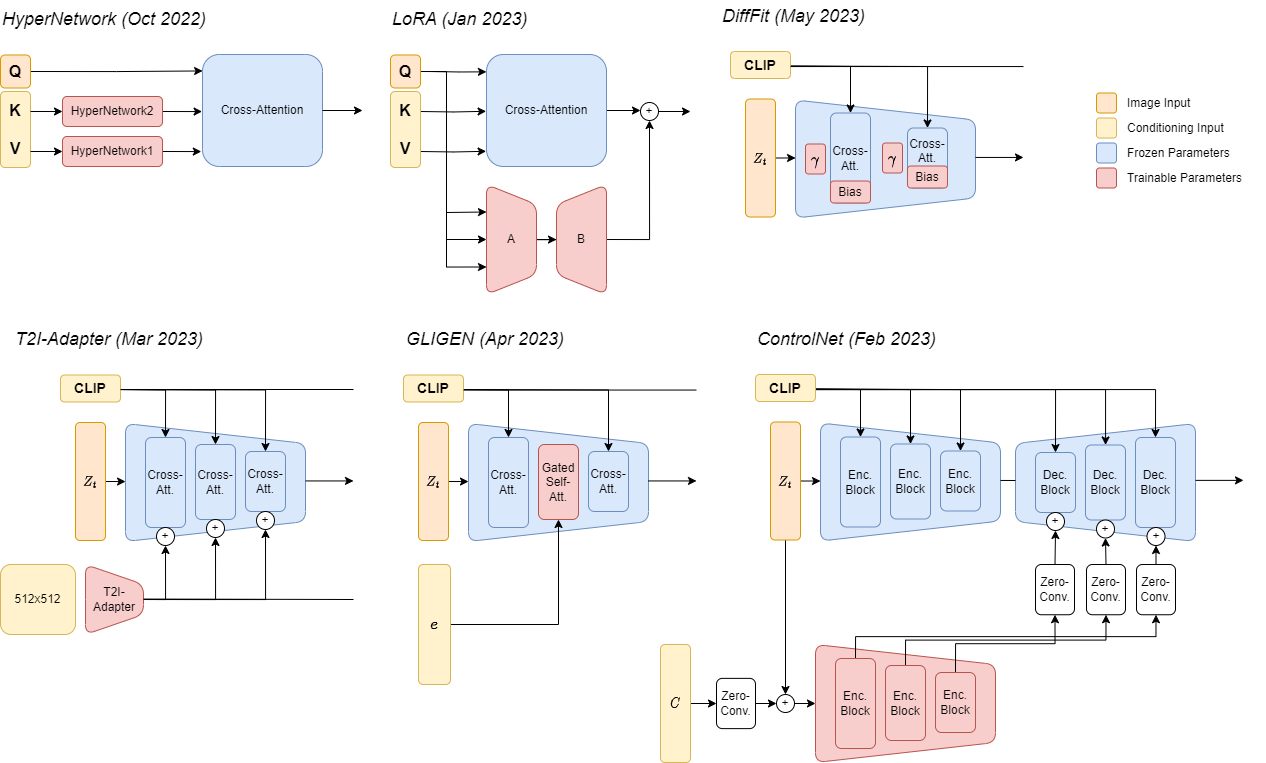}
\caption{Architectural overview of the mentioned "efficient fine-tuning" (top-row) and "conditional fine-tuning" (bottom row) techniques.}
\label{fig:techniqueoverview}
\end{figure*}

A graphical summarization of these techniques can be seen in \autoref{fig:techniqueoverview}.

\subsection{Image Generation from Scene Graphs}
Graphs offer a powerful means of representing various types of data, including citations, social and economic networks, biochemical and knowledge graphs, as they explicitly describe relationships and dependencies. Thus, several methods have been developed to learn from such graphs, known as Graph Neural Networks (GNN) \cite{gnn}.\\

Using text as conditioning in image generation models has certain drawbacks when compared to other approaches. Image descriptions may be long and loosely structured and the meaning depends strongly on syntax and language. Furthermore, text inherently carries ambiguity, as multiple sentences can express the same concept \cite{transformerbased}. Thus, in certain cases where image descriptions tend to be complex, scene graphs provide a better approach for conditioning image generation models. Additionally, text embedded by CLIP, which is the standard in current diffusion models, has some weaknesses on more systematic or complex tasks e.g. counting objects or distances between objects \cite{CLIP}. Moreover, CLIP is generally not good at capturing structural information e.g. CLIP scores are almost the same for "The dog is on the motorcycle" and "The motorcycle is on the dog" for an image corresponding to the first description \cite{structureclip}. GLIGEN \cite{gligen} uses bounding boxes to ground the text embeddings for more precise control of location of objects. However, A significant challenge in utilizing scene graphs for image generation lies in the complexity of aligning visual and graph features \cite{diffusionbased}. Most research on scene graphs relies on the Visual Genome dataset, which contains human-annotated scene graphs \cite{sg2im}.\\

The first image generation model that utilizes scene graphs, Sg2Im \cite{sg2im}, uses a Graph Convolutional Network (GCN) to obtain object features and scene layouts, which are a course image representations of the intended scenes. The scene layouts are then up-scaled and refined to generate realistic images. The generation process is trained adversarially, optimizing GAN losses. PasteGAN \cite{pastegan} conditions a GAN on image crops that contain single objects corresponding to certain nodes in a scene graph. Later, Sortino et al. \cite{transformerbased} used a similar approach to predict scene layouts from scene graphs and condition a VQVAE on those embeddings, thus avoiding the instability and mode collapse problems of GANs. Recently, Yang et al. \cite{diffusionbased} proposed a method to condition a latent diffusion model on scene graphs, by learning embeddings using masked contrastive pre-training to align image and graph embeddings. This is done by masking out certain parts of an image and reconstructing it using scene graph encodings. This encoder can then be utilized to embed scene graphs and feed the embeddings into a latent diffusion model using cross-attention. On April 2023, Farshad et al. \cite{scenegenie} (LMU CompVis, the latent diffusion group) presented a method which utilizes CLIP embeddings of nodes in a scene graph to predict scene layouts using a GCN. These scene layouts are then used as guidance in a latent diffusion model.

\section{Method}
To condition a latent diffusion model on scene graphs, we make use of two state-of-the-art conditioning techniques, ControlNet \cite{controlnet} and Gated Self-Attention \cite{gligen}. Two approaches that depend on the technique can be defined:
\begin{enumerate*}
  \item predict scene layouts from scene graphs
  \item directly predict scene embeddings from scene graphs.
\end{enumerate*}
Scene layouts or scene embeddings can then be used as conditioning. We use these techniques instead of fine-tuning a diffusion model like all previous works, because datasets of scene graphs are quite limited in size e.g. the largest scene graph dataset Visual Genome \cite{visualgenome} only consists of 108,077 data points, i.e. $5 \times 10^4$ times smaller than LAION5B.

\subsection{Graph Convolutional Network}
Graph Convolutional Networks (GCNs) \cite{gcn} are a type of neural network designed to work with graph-structured data. In contrast to traditional convolutional neural networks (CNNs) that process grid-like data e.g., images, GCNs can handle complex relationships present in graphs. The key idea behind GCNs is to generalize convolutional operations to graphs. In a graph, nodes represent entities, and edges represent connections between them. GCNs leverage this structural information to perform convolution-like operations. They aggregate and combine information from neighboring nodes to compute new node representations. This aggregation involves taking a weighted sum of the feature vectors of neighboring nodes and combining them with the node's own features. By stacking multiple graph convolutional layers, GCNs can capture deeper structural information.\\

We first embed each individual node and edge description using CLIP 
to obtain text embeddings $c$ and then apply multiple layers of graph convolution to compute graph features $f_g$. Because the meaning of each node is encoded separately using CLIP and defined explicitly in the graph, the resulting features are more robust and not limited by CLIP's lack of structural semantics.

\subsection{Via Scene Layouts}
\begin{figure*}[h]
\centering
\includegraphics[width=0.9\textwidth]{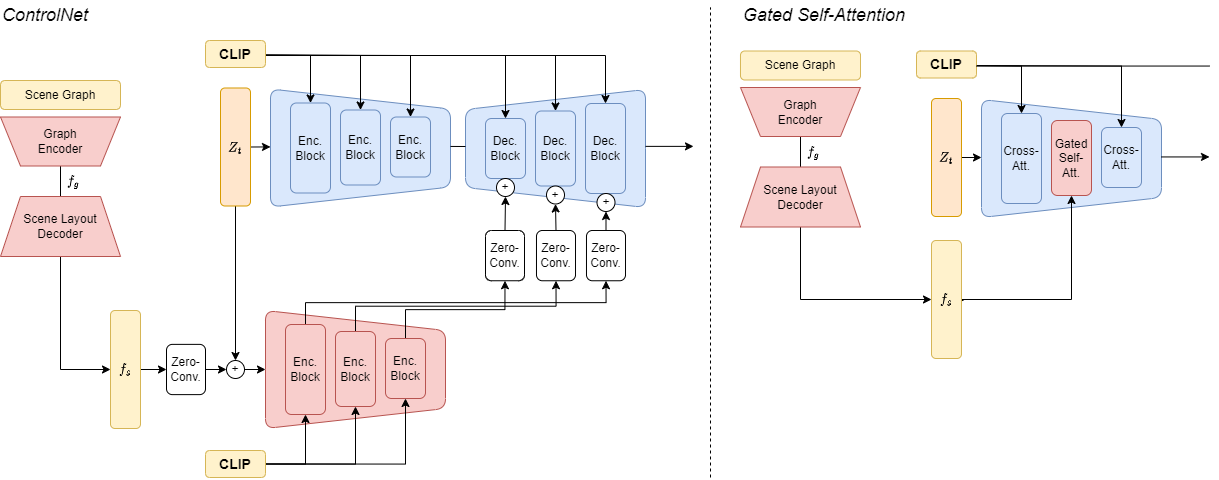}
\caption{Architectures of both approaches that utilize scene layouts as an intermediate step for image generation.}
\label{fig:viascenelayouts}
\end{figure*}

We use the graph encoder explained above to obtain graph features $f_g$, by jointly pretraining a small CNN decoder to predict course scene layouts $f_s$, similar to Johnson et al. \cite{sg2im} but using BERT \cite{bert} and CLIP \cite{CLIP} embeddings. For ControlNet, the dimensions of a scene layout has to be the same as $z_T$ i.e. $64 \times 64 \times 128$. Thus, we use the decoder only up until an intermediate layer and fine-tune it together with the graph encoder while training using ControlNet. In this approach, the text prompt, that is encoded with CLIP, is set to the default prompt "an image" during training. A default prompt is used prevent the CLIP text conditioning to influence the image generation process. An illustration of both approaches can be seen in \autoref{fig:viascenelayouts}.

The combination of GCN and CNN to transform graphs to an image representation is necessary in this approach, because the denoising U-Net that is fine-tuned using ControlNet expects image-like representations such as depth maps, semantic maps, edges or poses. During this transformation, spatial information like location and shape can be estimated from the object and relation nodes of the scene graph respectively.

\subsection{Via Graph Embeddings}
\begin{figure*}[h]
\centering
\includegraphics[width=0.73\textwidth]{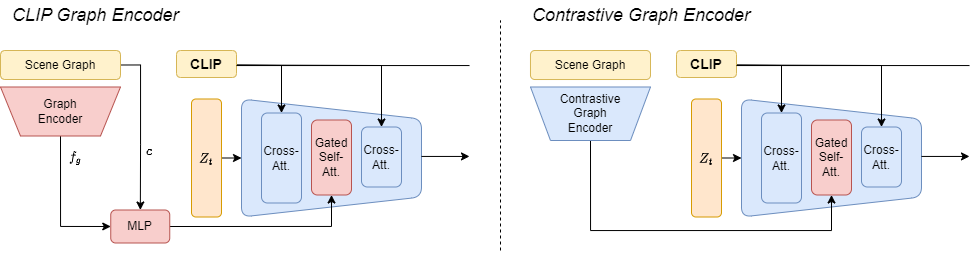}
\caption{Architectures of both approaches that utilize graph embeddings for image generation.}
\label{fig:viagraphembeddings}
\end{figure*}

The resulting graph features $f_g$ from the scene layout pretraining can also be directly used in Gated Self-Attention, without the need of encoding scene layouts. Controlling spatial arrangements using with ControlNet is rigid and therefore the structure of the input image is crucial. Gated Self-Attention allows for a more loosely control of spatial arrangements e.g. using bounding boxes, key points or other conditional signals. By creating scene layouts in the pre-training phase, information about specific words such as actions can get lost. This can be prevented, by fusing the text embeddings $c$ with the corresponding graph features $f_g$ using $h = MLP(concatenate(c, f_g))$ \cite{gligen}. Here, the CLIP encoder for the graph embedding can also be replaced by BERT \cite{bert}, similar to \cite{structureclip}. The default prompt is used and the graph encoder is trainable during fine-tuning, for architecture see \autoref{fig:viagraphembeddings}.\\

Possibly even better embeddings can be obtained by contrastively learning similarities between images and scene graphs, similar to CLIP but for graphs instead of text. Standard CLIP does not include negatives for structural relationships. Within a batch there are no counter-examples for captions such as "the large bus and the green grass" $\rightarrow$ "the green bus and the large grass". To do this a scene graph can be modified by switching attributes that are connected to different objects.

The concept was initially introduced in Structure-CLIP\cite{structureclip}, where scene graphs were utilized to mine hard negatives. Instead of fine-tuning CLIP on graphs, we leverage the frozen graph knowledge transformer from Structure-CLIP to obtain meaningful graph embeddings - see \autoref{fig:viagraphembeddings}.

\subsection{Structure-CLIP}
Since CLIP, as a contrastive learning method, does not provide negative samples for structural information, it performs poorly on tasks that require detailed semantics. Structure-CLIP \cite{structureclip} proposed a composition-aware hard negative mining strategy, which involves manipulating scene graphs in a meaningful way. That is, a sampling function that swaps nodes in a scene graph, to alter the caption in a way such that the sentence stays similar but its semantics change. Scene graphs are used because not all possible swaps of nodes change the semantics or form a correct sentence. Additionally, the authors propose a "knowledge-enhanced structured framework" to further capture structured knowledge from triples formed from the scene graph. A triple is a pair of nodes and their relation e.g. (Cows, Sit in, Hay). Each word in a triple is embedded using BERT \cite{bert} and each triple is then fed through a modified Transformer encoder \cite{AIAYN} with
\begin{equation*}
  \begin{aligned}
    Q=H W_Q, K&=H W_K, V=H W_V,\\
    Attention(H) &= softmax\left(\frac{Q K^{\top} \odot M}{\sqrt{d_K}}+(1-M) * \delta\right) V,
    \end{aligned}
\end{equation*}
where $H$ is the sequence of embedded triples, $\odot$ is element-wise multiplication, $M_{ij} = 0$ and $\delta << 0$, so the output is near $0$ at the beginning. Relying only on structured knowledge may lead to a loss of the original semantics, to this end the triple embedding is added to the CLIP text embedding. Incorporating the triple encoding transformer significantly improved the performance, compared to just using their negative mining strategy on its on.

\section{Experiments}
\subsection{Dataset}
For the fine-tuning dataset we choose Visual Genome \cite{visualgenome}, which is comprised of 108,077 scene graph and image pairs, with additional annotations such as bounding boxes and object attributes. For pretraining, we use COCO-Stuff \cite{cocostuff}, which contains 45,000 images with corresponding bounding boxes and segmentation masks, where we synthetically generate scene graphs, following Johnson et al. \cite{sg2im}.

\subsection{Settings}
As a base model we always use Stable Diffusion 2.1 \footnote{\url{https://huggingface.co/stabilityai/stable-diffusion-2-1}}. As a baseline we create text description from the scene graph and feed them into the base model as CLIP embeddings. We also fine-tune the base model on Structure-CLIP embeddings using LoRA \cite{lora}. For a second approach, we pretrain the scene layout model and use it to fine-tune the model using ControlNet. We then use the graph embeddings directly to fine-tune the model using Gated Self-Attention. We also compare different text embeddings, namely BERT \cite{bert} and CLIP \cite{CLIP}. Finally, we make use of the graph transformer from Huang et al. \cite{structureclip} to create the embeddings for the Gated Self-Attention model. We use FID-score for measuring the image sampling quality. The default prompt for all models except for the first two is "an image". To summarize, these are the experiments we conduct:
\begin{enumerate}[label=(\roman*)]
\item Text (CLIP) $\rightarrow$ Diffusion
\item Text (Structure-CLIP) $\rightarrow$ Diffusion (LoRA)
\item Scene Layouts (BERT) $\rightarrow$ ControlNet
\item Scene Layouts (CLIP) $\rightarrow$ ControlNet
\item Embedding (BERT) + Text $\rightarrow$ Gated Self-Attention
\item Embedding (CLIP) + Text $\rightarrow$ Gated Self-Attention
\item Embedding (Structure-CLIP) + Text $\rightarrow$ Gated Self-Attention
\end{enumerate}

\subsection{Results}
\begin{figure*}
\centering
\includegraphics[width=\textwidth]{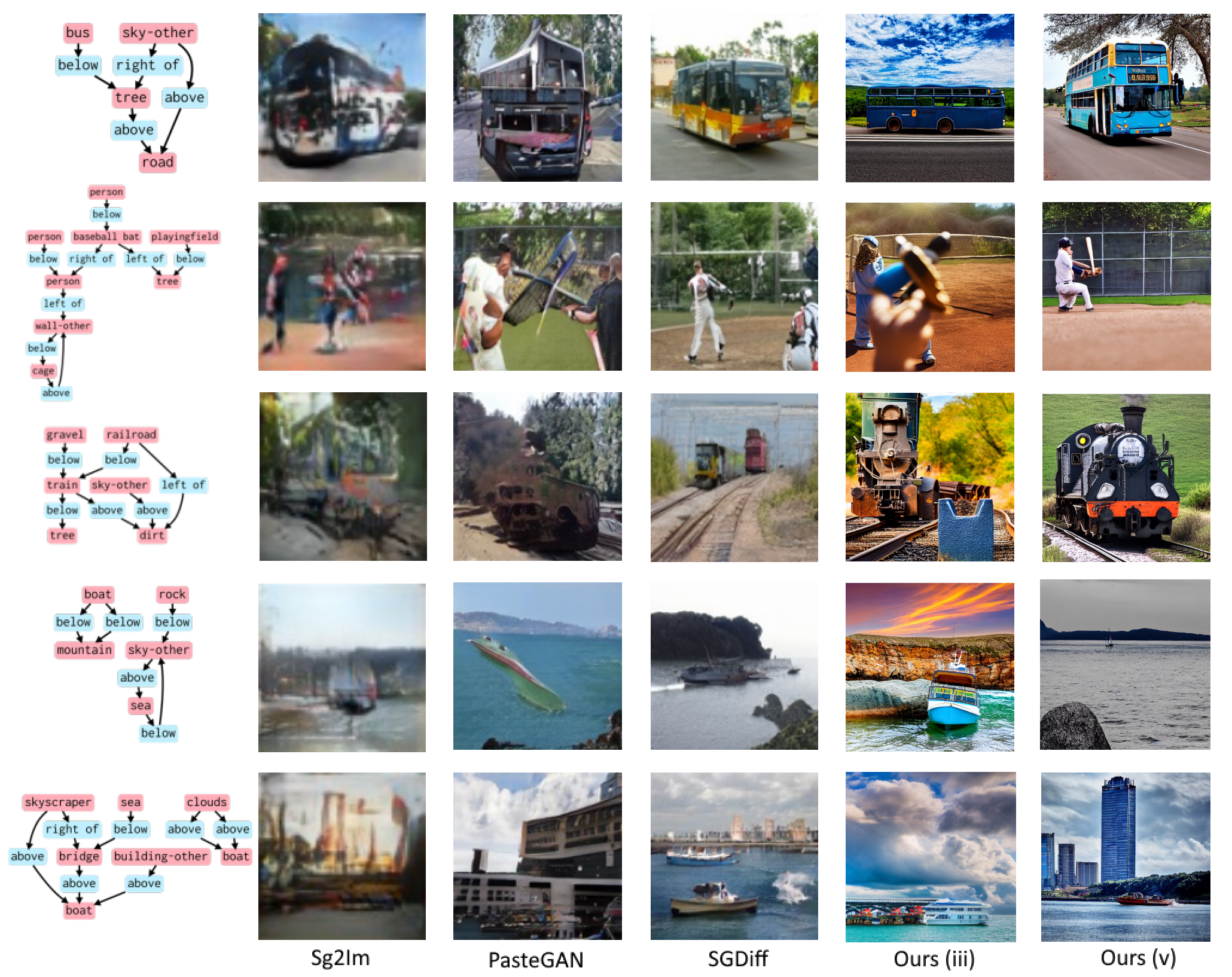}
\caption{Examples of different models from the same scene graphs.}
\label{fig:examples}
\end{figure*}

\begin{figure*}
\centering
\includegraphics[width=\textwidth]{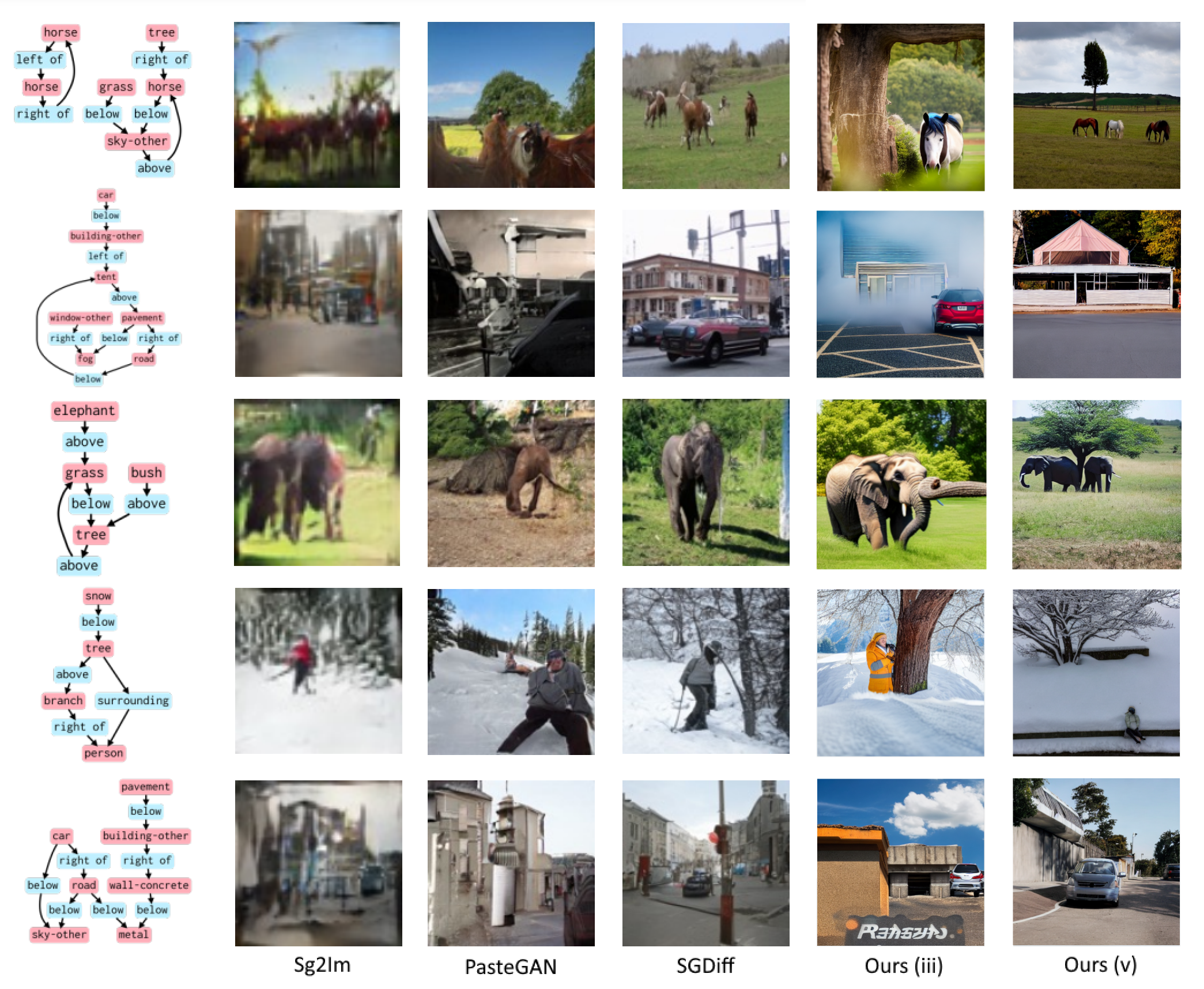}
\caption{Examples of different models from the same scene graphs.}
\label{fig:examples2}
\end{figure*}

The values of each model are either taken from the paper directly or from Yang et al. \cite{diffusionbased}. Each model is evaluated without using the ground truth masks. Some examples can be seen in \autoref{fig:examples} and \autoref{fig:examples2}. The results on COCO Stuff \cite{cocostuff} and Visual Genome \cite{visualgenome} can be seen in \autoref{tbl:results}.\\

\begin{figure}[h]
\renewcommand{\arraystretch}{1.5}
\begin{tabular}{lcccc}
\hline
\textbf{Method} & \multicolumn{2}{c}{\textbf{Inception Score $\uparrow$}} & \multicolumn{2}{c}{\textbf{FID $\downarrow$}} \\
\cline{2-5}
& \textbf{COCO} & \textbf{VG} & \textbf{COCO} & \textbf{VG} \\
\hline
Real Image (256x256) & 30.7 & 27.3 & - & - \\
Sg2Im \cite{sg2im} & 8.2 & 7.9 & 99.1 & 90.5 \\
PasteGAN \cite{pastegan} & 12.3 & 8.1 & 79.1 & 66.5 \\
SGTransformer \cite{transformerbased} & 13.7 & 12.8 & 52.3 & 60.3 \\
SGDiff \cite{diffusionbased} & 17.8 & 16.4 & \textbf{36.2} & 26.0 \\
SceneGenie \cite{scenegenie} & 22.2 & 20.3 & 63.3 & 42.2 \\
Ours (iii) \textsuperscript{\ref{note:result}} & 32.1 & - & 279.3 \textsuperscript{\ref{note:result}} & - \\
Ours (v) \textsuperscript{\ref{note:fid}} & \textbf{32.2} & - & 266.4 \textsuperscript{\ref{note:fid}} & - \\
\hline
\end{tabular}
\caption{Results of our experiment compared to several other methods.}
\label{tbl:results}
\end{figure}

\footnotetext[2]{\label{note:result}Because of time and resource constraints, we were unable to train or fine-tune any model. Additionally, SGTransformer, SGDiff and Structure-CLIP are not published yet. Therefore, we use the pre-trained scene layout network from sg2im to perform two experiments:
\begin{enumerate*}
\item Scene Layouts (bounding boxes, sg2im) $\rightarrow$ GLIGEN
\item Scene Layouts (semantic map, sg2im) $\rightarrow$ ControlNet
\end{enumerate*}
with 10 scene graphs from COCO for each experiment.
}
\footnotetext[3]{\label{note:fid}FID requires around 50k images according to its authors. Since we only provide 10 images, these numbers are probably not meaningful.}

\section{Discussion}
The proposed approaches for fine-tuning diffusion models using scene graphs have demonstrated promising results and significant advancements in image generation. By incorporating scene graphs as a more structured and precise representation of image content, we have achieved improved control while maintaining high image quality. However, there are several important points to consider and discuss.
Because of time and resource constraints, we were not able to train any model. Thus, the overall image quality and control can be significantly improved. Scene layouts, including bounding boxes produced by sg2im \cite{sg2im} are not optimal and very course, which is the current bottleneck of our system. The pre-trained models used in our experiments were not fine-tuned to the input, which results in missing image attributes such as objects or relations or overall bad image quality.

\section{Conclusion}
In conclusion, diffusion models have emerged as powerful image generation models, but conditioning on text for image generation has limitations due to the ambiguity and length of textual descriptions. However, scene graphs offer a structured and unambiguous representation of image content as they provide a precise encoding of object relationships and spatial arrangements, making them a superior choice for image generation when fine-grained control and accurate synthesis are crucial. Different methods have been proposed for utilizing scene graphs in image generation, including predicting scene layouts, conditioning on object crops corresponding to nodes in a scene graph, and learning embeddings through masked contrastive pre-training. However, images conditioned on scene graphs are not yet comparable to images conditioned on other modalities like bounding boxes, poses, or depth maps.

In our work, we have contributed multiple approaches for effectively fine-tuning diffusion models using scene graphs. We have evaluated these methods to some extent and provided the code for further research and exploration in this area. Our contributions aim to enhance the capabilities of diffusion models and facilitate more precise and controlled image generation. Overall, the advancements in diffusion models and the utilization of scene graphs offer exciting prospects for the field of image generation. Further research and development in this area can lead to even more impressive results and advancements in generating images with fine-grained control and accuracy.

\acknowledgements{
I want to thank Ahi's Pizza who makes the best pizza in Ulm and whoever invented matcha.}

\bibliographystyle{abbrv}
\bibliography{bibliography}

\end{document}